\documentclass{sig-alternate}
\pdfoutput=1
\usepackage{epsfig}
\usepackage{amssymb}
\usepackage{amsmath}
\usepackage{amsfonts}
\usepackage{multirow}
\usepackage{url}
\usepackage[numbers]{natbib}
\usepackage{microtype}
\usepackage{booktabs}
\usepackage{subcaption}
\usepackage{times}
\usepackage[usenames,dvipsnames,svgnames,table]{xcolor}
\usepackage{tikz}
\usetikzlibrary{calc,decorations.pathreplacing,matrix}
\DeclareMathOperator*{\argmax}{arg\,max}

\newcommand{\entA}{\ensuremath{{S_A}}}
\newcommand{\entC}{\ensuremath{{S_C}}}
\newcommand{\entB}{\ensuremath{{S_B}}}
\newcommand{\entP}{\ensuremath{{S_P}}}
\newcommand{\entU}{\ensuremath{{S_U}}}
\newcommand{\entW}{\ensuremath{{S_W}}}

\newcommand{\relA}{\ensuremath{A} }
\newcommand{\relC}{\ensuremath{C} }
\newcommand{\relR}{\ensuremath{R} }
\newcommand{\relUW}{\ensuremath{UW} }
\newcommand{\relBW}{\ensuremath{BW} }
\newcommand{\relPW}{\ensuremath{PW} }

\newcommand{\cut}[1]{}

\newcommand{\para}[1]{\noindent\textbf{#1}} 
\newcommand{\highest}[1]{\textbf{#1}} 

\begin{document}
%

\title{Collectively Embedding Multi-Relational Data\\for Predicting User Preferences}

\numberofauthors{2} 
%
\author{
%
%
\alignauthor
Nitish Gupta\\
       \affaddr{Indian Institute of Technology}\\
       \affaddr{Kanpur, India}\\
       \email{nitishgupta.iitk@gmail.com}
\alignauthor
Sameer Singh\\
       \affaddr{University of Washington}\\
       \affaddr{Seattle, WA}\\
       \email{sameer@cs.washington.edu}
}

\date{12 February, 2015}

\maketitle
\begin{abstract}
Matrix factorization has found incredible success and widespread application as a collaborative filtering based approach to recommendations.
Unfortunately, incorporating additional sources of evidence, especially ones that are incomplete and noisy, is quite difficult to achieve in such models, however, is often crucial for obtaining further gains in accuracy.
For example, additional information about businesses from reviews, categories, and attributes should be leveraged for predicting user preferences, even though this information is often inaccurate and partially-observed.
Instead of creating customized methods that are specific to each type of evidences, in this paper we present a generic approach to factorization of relational data that collectively models all the relations in the database.
By learning a set of embeddings that are shared across all the relations, the model is able to incorporate observed information from all the relations, while also predicting all the relations of interest.
Our evaluation on multiple Amazon and Yelp datasets demonstrates effective utilization of additional information for held-out preference prediction, but further, we present accurate models even for the \emph{cold-starting} businesses and products for which we do not observe \emph{any} ratings or reviews.
We also illustrate the capability of the model in imputing missing information and jointly visualizing words, categories, and attribute factors. 
\end{abstract}




\section{Introduction}

Predicting user preferences, for items such as for commercial products, movies, and businesses, is an important and well-studied problem in recommendation systems.
Collaborative filtering using matrix factorization~\cite{koren2009matrix}, in particular, has found widespread adoption as the tool of choice for this problem.
By relying on co-occurrences in the ratings, however, these methods do not perform well on users or items that do not have ample observed ratings, i.e. users and items that are rare or new to the system.


Fortunately, since users and items are part of a larger database, extra relational information about such users and items can often be utilized for predicting preferences.
It is, for example, often not difficult to obtain information such as product categories, album genres, review text, and attributes/features of the items, however this external evidence is rarely complete or noise-free. 
A number of existing approaches have thus been proposed to use these sources of information for improving user preferences.
\citet{koren08kdd}, for example, combines the factorization model with an encoding of the external information as fully-observed features. 
Several studies have also investigated algorithms for incorporating specific sources of information, for example modeling user reviews~\cite{mcauley2012learning, ling2014ratings, ganu2009beyond},  integrating context information \cite{karatzoglou2010multiverse, hariri2014context}, exploiting item taxonomy \cite{koenigstein2011yahoo, weng2008exploiting} and learning changes in user preferences and expertise over time \cite{koren2010collaborative, mcauley2013amateurs} to improve rating prediction.
However, these approaches face a number of disadvantages when applied to heterogeneous, incomplete, multi-relational schema common in practice.
First, these approaches are designed for certain \emph{types} of relations and are restricted to relations of that type.
Thus, it is not clear how additional sources of information can be incorporated, for example, how partially observed product categories can be used to improve rating prediction in \citet{mcauley2012learning}. 
Further, by training the model to predict entries of only one or two relations, these approaches ignore the dependencies between other relations and entities in the database, such as simultaneously predicting the cuisine of a restaurant, and the users that will like it, from the user reviews of the restaurant.
There's a need for a generic machine learning approach that is able to leverage the dependencies between users, items, and additional data for estimating user preferences more accurately. 


In this paper, we present a collective factorization model for incorporating heterogeneous relational data for user preference prediction in a domain-independent manner. 
Collective factorization assigns a latent low-dimensional vector (an \emph{embedding} or \emph{factor}) for every entity in the database that is used to predict all of the observed relations between pairs of entities. 
The collective model thus extends the intuition behind matrix factorization based recommendation systems that include embeddings for every user and business/product, and is a generalization of \citep{mcauley2012learning} that assign factors to every user, business/product, and review words.
Since the latent embeddings in collective factorization are used to model all of the observed entries in the database, it is capable of predicting \emph{any} type of relation between entities.
Training the embeddings to capture all of the dependencies also makes it easy to integrate multiple evidences for the same relation; incorporating another source of information is as simple as including an additional relation/table in the database.
Further, since the embeddings for all the entities are defined over the same low-dimensional space, we can compute similarity between any pair of entities, even if they are not directly observed in the same relation.
The collective factorization model provides further benefits for practical deployment: the training algorithm is efficient and scalable, and the model complexity can be controlled by varying the embedding dimensionality.

We present a four-way evaluation of the collective factorization model (\S\ref{sec:model}), as applied to the Yelp\footnote{\texttt{www.yelp.com/dataset\_challenge}} and Amazon datasets\footnote{\texttt{www.amazon.com}} (\S\ref{sec:cmf-for-db}).
(1)~We demonstrate that the collective factorization model is effective in incorporating additional sources of information in \S\ref{sec:results:held-out}, in particular provides significant accuracy gains for predicting user preferences. 
(2)~In \S\ref{sec:results:cold-start}, we show that the proposed model is especially useful for \emph{cold-start} estimation, e.g. for estimating preferences for \emph{new} businesses and products for which no reviews or ratings have been observed. 
(3)~An advantage of the model is that it can be used to impute missing values in the external data; we present an evaluation of this capability in \S\ref{sec:results:imputing-db}.
(4)~We explore the implicit relations learned by the model that were not observed in the data by visualizing categories, business attributes, and the review words in the same two-dimensional plot in \S\ref{sec:results:qualitative}. 

\section{Collective Factorization}
\label{sec:model}

In this section, we present the probabilistic collective matrix factorization that jointly models the relations between entities, by leveraging data from all the other relations the entities participate in.

\subsection{Relational Data}

We represent relational data as a set of entities ($\mathcal{E}$) and relations between them ($\mathcal{R}$).
Formally, the observed database, denoted by $\mathcal{D}$, consists of tuples of the form $\{r_t,e_{t1},e_{t2},y_t\}_{t=1}^{T}$, where $r_t\in\mathcal{R}$ is a relation, $e_{t1},e_{t2}\in\mathcal{E}$ are a pair of entities, and $y_t\in\{0,1\}$ denotes whether $r_t(e_{t1},e_{t2})$ holds (or not).
For example, a simple database that consists only of the user preferences would contain products and users as the entities, and only a single relation $r$, such that $r(e_{t1},e_{t2})=1$ if user $e_{t1}$ liked the product $e_{t2}$.
As is clear from this example, many databases in real-life are only sparsely observed, in that only a very small set of possible relations are observed, and the goal of modeling such datasets is to be able to \emph{complete} this database.
Specifically, given any query $r_q(e_{q1},e_{q2})$ that is absent from the observed database, we would like to predict whether the relation holds.
Further, as we will see later, user's preferences for items can be represented by one of the relations, with additional information about users and items as other relations. 

\subsection{Collective Factorization Model}

Collective matrix factorization model~\cite{singh2008relational} extends the commonly-used matrix factorization model to multiple matrices by assigning each entity a low-dimensional latent vector that is shared across all the relations the entity appears in.
Formally, we assign each entity $e$ in our database a $k$-dimensional embedding vector (latent factor) $\phi_{e} \in \mathbb{R}^{k}$ (the set of these embeddings for all the entities in the database is $\Phi$). 
We model the probability that $r(e_1,e_2)$ holds by:
\begin{equation}
P_\Phi\left[r(e_1,e_2)=1\right] = \sigma(\phi_{e1}\cdot\phi_{e1})\label{eq:model}
\end{equation}
where $\sigma$ is the \emph{sigmoid} function, $\sigma(s)=\frac{1}{1 + e^{-s}}$\footnote{As is common in recommendation systems, we can also include per entity bias and a global matrix offset to Eq.~\ref{eq:model}. We skip showing the biases in our formulation for brevity.}.
The probability that $r(e_1,e_2)=y$ is,
\begin{equation}
P_\Phi\left[r(e_1,e_2)=y\right] = \sigma(\phi_{e1}\cdot\phi_{e1})^y(1 - \sigma(\phi_{e1}\cdot\phi_{e1}))^{(1 - y)}
\end{equation}

The collective factorization model presents a number of advantages. 
By sharing the entity embeddings amongst all the relations, it is able to capture all the sources of evidence in a joint manner, for example the embeddings used to predict preferences for a user will leverage information from other users' preferences in a collaborative filtering fashion, but also from business attributes, categories, and words that appear in the reviews (the details of the model as applied to the datasets are described in \S\ref{sec:cmf-for-db}). 
The sharing of embeddings also allows them to be used to predict any of the relations in the database, i.e. along with predicting ratings, we can also predict business categories, attributes, and the text of the reviews.
A further advantage of learning collective embeddings is that all the entities are effectively \emph{embedded} in the same $k$-dimensional space, and thus similarities and distances can be computed and analyzed for any set of entities (we explore such visualizations in \S~\ref{sec:results:qualitative}).
Finally, test-time inference takes constant time and thus is incredibly efficient: we only need a dot-product between low-dimensional vectors for estimating the probability of a relation to hold between a pair of entities.

\subsection{Estimating Entity Factors}
To estimate the parameters i.e. latent vectors $\Phi$, we maximize the regularized log likelihood of the observed training instances (observed entries in the database, $\mathcal{D}$). Specifically, we maximize:

\begin{align}
\hat{\Phi} &=  \argmax_{\Phi}~l(\mathcal{D},\Phi)\\
l(\mathcal{D},\Phi)&= \sum_{t=1}^{T} \log P_\Phi\left[r_t(e_{t1},e_{t2})=y_t\right] - \lambda\left(\|\Phi\|_{2}^{2}\right)
\label{likelihood}
\end{align}

To optimize this objective and estimate the latent factors, we use stochastic gradient descent (SGD) by cycling over the entries of the database multiple times, updating the latent factors in the direction of stochastic gradient for each entry.
In particular, the $i^\text{th}$ update that uses $t^\text{th}$ database entry is given by, 
\begin{align}
\phi_{e_{t1}}^{(i+1)} &\leftarrow \phi_{e_{t1}}^{(i)} + \gamma \left(e_{t} * \phi_{e_{t2}}^{(i)} - \lambda \phi_{e_{t1}}^{(i)}\right) \\
\phi_{e_{t2}}^{(i+1)} &\leftarrow \phi_{e_{t2}}^{(i)} + \gamma \left(e_{t} * \phi_{e_{t1}}^{(i)} - \lambda \phi_{e_{t2}}^{(i)}\right)
\end{align}
where $e_{t} = y_{t} - \sigma\left(\phi_{e_{t1}}^{(i)}\cdot\phi_{e_{t2}}^{(i)}\right)$ and $\gamma$ is the learning rate. Similarly, the update rules the for per entity bias and global matrix offset, if used, can be derived easily.
Along with strong theoretical properties and widespread empirical success, the algorithm is also memory and time efficient since it runs on a single entry at a time, and additionally, further potential for scalability via parallelism~\cite{Niu11hogwild,agarwal11} has been demonstrated in recent approaches.

\begin{figure*}[tb]
\centering
\hskip -20mm
\begin{tikzpicture}[scale=1.0, transform shape]
\newcommand{\at}[3]{
  \begin{scope}[shift={(#1,#2)}]
    #3
  \end{scope} 
}

\newcommand{\embcolor}{PineGreen}
\newcommand{\ye}{1}
\newcommand{\he}{1.5}
\newcommand{\we}{1}
\newcommand{\gape}{1}

\newcommand{\xc}{0.5}
\newcommand{\xa}{\xc+\we+\gape}
\newcommand{\xb}{\xa+\we+\gape}
\newcommand{\xu}{\xb+\we+\gape}
\newcommand{\xw}{\xu+\we+\gape}

\draw [thick]
	(\xb+\we/2,\ye+\he+0.5) node[above,\embcolor] {Entities, $\mathcal{E}$};
\draw [\embcolor,decorate,decoration={brace,amplitude=5pt},xshift=-15pt] 
	(\xc,\ye)  -- (\xc,\ye+\he) node [align=right,left,\embcolor,midway,xshift=-5pt]
	{Model Parameters, $\Phi$\\{\small $k$-dimensional entity vectors}};
\draw[<->,\embcolor!50]
	(\xu,\ye-0.15) -- (\xu+\we,\ye-0.15) 
	node[pos=.5,shape=rectangle,fill=white,draw=none] {\scriptsize $k$};
\draw[<->,\embcolor!50] 
	(\xu+\we+0.3,\ye) -- (\xu+\we+0.3,\ye+\he) 
	node[pos=.5,shape=rectangle,fill=white,draw=none] {\scriptsize $|\entU|$};
\draw [semithick,fill=gray!10] 
	(\xu,\ye) rectangle (\xu+\we,\ye+\he)
	(\xu+\we/2.0,\ye+\he/2.0) node[draw=none,shape=circle] (nodeU) {$\Phi_\entU$}
	(\xu+\we/2.0,\ye+\he) node[above] {\scriptsize Users, \entU};
\draw[<->,\embcolor!50]
	(\xb,\ye-0.15) -- (\xb+\we,\ye-0.15) 
	node[pos=.5,shape=rectangle,fill=white,draw=none] {\scriptsize $k$};
\draw[<->,\embcolor!50] 
	(\xb+\we+0.3,\ye) -- (\xb+\we+0.3,\ye+\he) 
	node[pos=.5,shape=rectangle,fill=white,draw=none] {\scriptsize $|\entB|$};
\draw [semithick,fill=gray!10] 
	(\xb,\ye) rectangle (\xb+\we,\ye+\he)
	(\xb+\we/2.0,\ye+\he/2.0) node[draw=none,shape=circle] (nodeB) {$\Phi_\entB$}
	(\xb+\we/2.0,\ye+\he) node[above] {\scriptsize Businesses, \entB};
\draw[<->,\embcolor!50]
	(\xc,\ye-0.15) -- (\xc+\we,\ye-0.15) 
	node[pos=.5,shape=rectangle,fill=white,draw=none] {\scriptsize $k$};
\draw[<->,\embcolor!50] 
	(\xc+\we+0.3,\ye) -- (\xc+\we+0.3,\ye+\he) 
	node[pos=.5,shape=rectangle,fill=white,draw=none] {\scriptsize $|\entC|$};
\draw [semithick,fill=gray!10] 
	(\xc,\ye) rectangle (\xc+\we,\ye+\he)
	(\xc+\we/2.0,\ye+\he/2.0) node[draw=none,shape=circle] (nodeC) {$\Phi_\entC$}
	(\xc+\we/2.0,\ye+\he) node[above] {\scriptsize Categories, \entC};
\draw[<->,\embcolor!50]
	(\xa,\ye-0.15) -- (\xa+\we,\ye-0.15) 
	node[pos=.5,shape=rectangle,fill=white,draw=none] {\scriptsize $k$};
\draw[<->,\embcolor!50] 
	(\xa+\we+0.3,\ye) -- (\xa+\we+0.3,\ye+\he) 
	node[pos=.5,shape=rectangle,fill=white,draw=none] {\scriptsize $|\entA|$};
\draw [semithick,fill=gray!10] 
	(\xa,\ye) rectangle (\xa+\we,\ye+\he)
	(\xa+\we/2.0,\ye+\he/2.0) node[draw=none,shape=circle] (nodeA) {$\Phi_\entA$}
	(\xa+\we/2.0,\ye+\he) node[above] {\scriptsize Attributes, \entA};
\draw[<->,\embcolor!50]
	(\xw,\ye-0.15) -- (\xw+\we,\ye-0.15) 
	node[pos=.5,shape=rectangle,fill=white,draw=none] {\scriptsize $k$};
\draw[<->,\embcolor!50] 
	(\xw+\we+0.3,\ye) -- (\xw+\we+0.3,\ye+\he) 
	node[pos=.5,shape=rectangle,fill=white,draw=none] {\scriptsize $|\entW|$};
\draw [semithick,fill=gray!10] 
	(\xw,\ye) rectangle (\xw+\we,\ye+\he)
	(\xw+\we/2.0,\ye+\he/2.0) node[draw=none,shape=circle] (nodeW) {$\Phi_\entW$}
	(\xw+\we/2.0,\ye+\he) node[above] {\scriptsize Review words, \entW};

\newcommand{\relcolor}{RoyalBlue}
\newcommand{\ym}{-2}
\newcommand{\hm}{2}
\newcommand{\wm}{2}
\newcommand{\gapm}{0.75}
\newcommand{\gridmcolor}{gray!10}

\newcommand{\xCB}{-2}
\newcommand{\xAB}{\xCB+\wm+\gapm}
\newcommand{\xR}{\xAB+\wm+\gapm}
\newcommand{\xBW}{\xR+\wm+\gapm}
\newcommand{\xUW}{\xBW+\wm+\gapm}

\newcommand{\sparsegrid}{
	\draw [step=0.1,thin,\gridmcolor] (0,0) grid (\wm,\hm);
	\foreach \i in {0,1,...,75}{
		\pgfmathsetmacro{\xcell}{int(rnd*20)*\wm/20}
		\pgfmathsetmacro{\ycell}{int(rnd*20)*\wm/20}
		\fill[gray!20] (\xcell,\ycell) rectangle (\xcell+0.1,\ycell+0.1);
	}
}

\draw [thick]
	(\xR+\wm/2,\ym-0.5) node[below,\relcolor] {Relations, $\mathcal{R}$};
\draw [\relcolor,decorate,decoration={brace,amplitude=5pt},xshift=-15pt] 
	(\xCB,\ym)  -- (\xCB,\ym+\hm) node [align=right,left,\relcolor,midway,xshift=-5pt]
	{Partial Observations\\{\small Predict missing data}};

\draw[<->,\relcolor!50]
	(\xCB,\ym+\hm+0.15) -- (\xCB+\wm,\ym+\hm+0.15) 
	node[pos=.5,shape=rectangle,fill=white,draw=none] {\scriptsize $|\entC|$};
\draw[<->,\relcolor!50] 
	(\xCB+\wm+0.3,\ym) -- (\xCB+\wm+0.3,\ym+\hm) 
	node[pos=.5,shape=rectangle,fill=white,draw=none] {\scriptsize $|\entB|$};
\at{\xCB}{\ym}{\sparsegrid}
\draw [thick] 
	(\xCB,\ym) rectangle (\xCB+\wm,\ym+\hm)
	(\xCB+\wm/2.0,\ym+\hm/2.0) node[draw=none,shape=circle] (nodeCB) {\relC}
	(\xCB+\wm/2.0,\ym) node[below] {\scriptsize Business Categories};
\draw[<->,\relcolor!50]
	(\xAB,\ym+\hm+0.15) -- (\xAB+\wm,\ym+\hm+0.15) 
	node[pos=.5,shape=rectangle,fill=white,draw=none] {\scriptsize $|\entA|$};
\draw[<->,\relcolor!50] 
	(\xAB+\wm+0.3,\ym) -- (\xAB+\wm+0.3,\ym+\hm) 
	node[pos=.5,shape=rectangle,fill=white,draw=none] {\scriptsize $|\entB|$};
\at{\xAB}{\ym}{\sparsegrid}
\draw [thick] 
	(\xAB,\ym) rectangle (\xAB+\wm,\ym+\hm)
	(\xAB+\wm/2.0,\ym+\hm/2.0) node[draw=none,shape=circle] (nodeAB) {\relA}
	(\xAB+\wm/2.0,\ym) node[below] {\scriptsize Business Attributes};
\draw[<->,\relcolor!50]
	(\xR,\ym+\hm+0.15) -- (\xR+\wm,\ym+\hm+0.15) 
	node[pos=.5,shape=rectangle,fill=white,draw=none] {\scriptsize $|\entU|$};
\draw[<->,\relcolor!50] 
	(\xR+\wm+0.3,\ym) -- (\xR+\wm+0.3,\ym+\hm) 
	node[pos=.5,shape=rectangle,fill=white,draw=none] {\scriptsize $|\entB|$};
\at{\xR}{\ym}{\sparsegrid}
\draw [thick] 
	(\xR,\ym) rectangle (\xR+\wm,\ym+\hm)
	(\xR+\wm/2.0,\ym+\hm/2.0) node[draw=none,shape=circle] (nodeR) {\relR}
	(\xR+\wm/2.0,\ym) node[below] {\scriptsize User/Business Ratings};
\draw[<->,\relcolor!50]
	(\xBW,\ym+\hm+0.15) -- (\xBW+\wm,\ym+\hm+0.15) 
	node[pos=.5,shape=rectangle,fill=white,draw=none] {\scriptsize $|\entW|$};
\draw[<->,\relcolor!50] 
	(\xBW+\wm+0.3,\ym) -- (\xBW+\wm+0.3,\ym+\hm) 
	node[pos=.5,shape=rectangle,fill=white,draw=none] {\scriptsize $|\entB|$};
\at{\xBW}{\ym}{\sparsegrid}
\draw [thick] 
	(\xBW,\ym) rectangle (\xBW+\wm,\ym+\hm)
	(\xBW+\wm/2.0,\ym+\hm/2.0) node[draw=none,shape=circle] (nodeBW) {\relBW}
	(\xBW+\wm/2.0,\ym) node[below] {\scriptsize Reviews for Business};
\draw[<->,\relcolor!50]
	(\xUW,\ym+\hm+0.15) -- (\xUW+\wm,\ym+\hm+0.15) 
	node[pos=.5,shape=rectangle,fill=white,draw=none] {\scriptsize $|\entW|$};
\draw[<->,\relcolor!50] 
	(\xUW+\wm+0.3,\ym) -- (\xUW+\wm+0.3,\ym+\hm) 
	node[pos=.5,shape=rectangle,fill=white,draw=none] {\scriptsize $|\entU|$};
\at{\xUW}{\ym}{\sparsegrid}
\draw [thick] 
	(\xUW,\ym) rectangle (\xUW+\wm,\ym+\hm)
	(\xUW+\wm/2.0,\ym+\hm/2.0) node[draw=none,shape=circle] (nodeUW) {\relUW}
	(\xUW+\wm/2.0,\ym) node[below] {\scriptsize Reviews by Users};

\newcommand{\arrowcolor}{black}
\draw[\arrowcolor,->] (nodeC) -- (nodeCB);
\draw[\arrowcolor,->] (nodeB) -- (nodeCB);

\draw[\arrowcolor,->] (nodeA) -- (nodeAB);
\draw[\arrowcolor,->] (nodeB) -- (nodeAB);

\draw[\arrowcolor,->] (nodeU) -- (nodeR);
\draw[\arrowcolor,->] (nodeB) -- (nodeR);

\draw[\arrowcolor,->] (nodeW) -- (nodeBW);
\draw[\arrowcolor,->] (nodeB) -- (nodeBW);

\draw[\arrowcolor,->] (nodeU) -- (nodeUW);
\draw[\arrowcolor,->] (nodeW) -- (nodeUW);
\end{tikzpicture}
\caption{\label{cmf-yelp}{\bf Collective Factorization for the Yelp Dataset:} Overview of the entities and the relations, with the latter represented by sparsely-observed matrices. The collective factorization model contains low-dimensional dense embeddings for all the entities which are used to model the respective relations the entities appear in (denoted by arrows). Collective factorization for Amazon is similar, the difference being that instead of businesses, Amazon contains \emph{products} and does not contain \emph{attribute} data.}
\end{figure*}
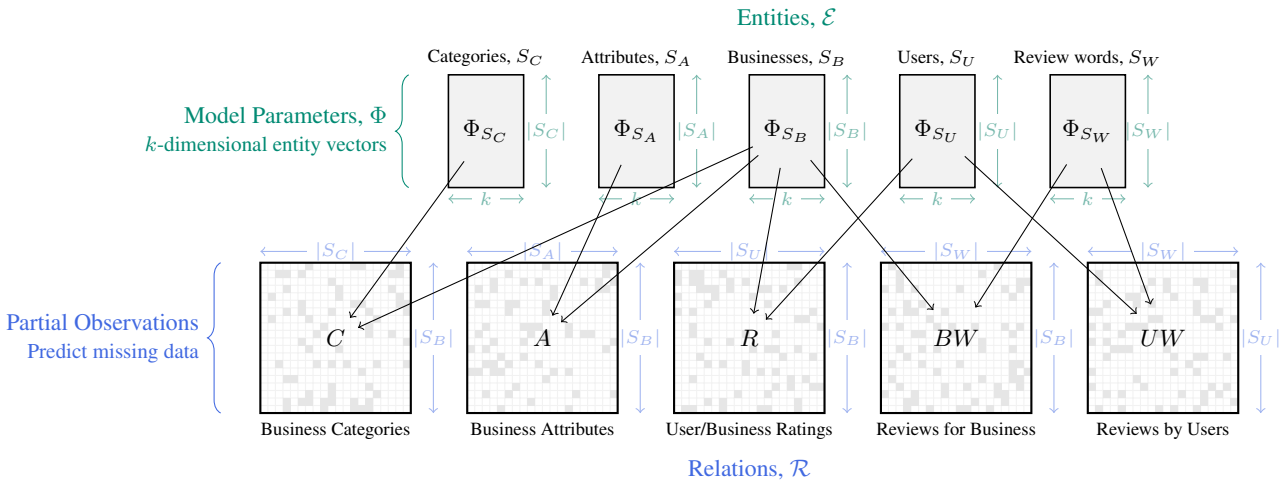

\section{Collective Factorization for\\Preference Prediction}
\label{sec:cmf-for-db}

The primary goal of recommendation systems is to provide personalized recommendation of services and products to users, often by learning their preferences from past rating history.
However, additional information can be leveraged to further improve the predictions, especially for users and items that are infrequent or new to the system.
For example, ratings are often accompanied by reviews whose text is undoubtedly informative about a user's tastes. 
The prevalence of such reviews has lead to tailored approaches of combining text reviews with rating history to improve the rating predicting task. 
\citet{ganu2009beyond}, for instance, alleviate the problem of coarse star ratings not being expressive by predicting topic based ratings solely from the text reviews. 
\citet{mcauley2012learning} integrate text reviews with star ratings by aligning the item \emph{latent factors} with the review topic vectors obtained from topic models in order to learn better embeddings for items, based on the topics users write about in their reviews.
Although such approaches are able to model text accurately, it is unclear how they generalize to other forms of additional information such as structured data about products/businesses and users in the form of business attributes, product categories, users' social networks, and so on. 
This rich and diverse data can clearly aid in user modeling, however existing approaches that are domain-specific or assume data to be fully-observed, do not generalize to this richly heterogeneous and noisy data. 
%
In the following section we describe how the \emph{collective matrix factorization} described in \S\ref{sec:model}, can be used to collectively model all kinds of relations in such databases to more accurately predict user preferences, and at the same time, help in completing the missing information in the database.

\subsection{Capabilities of Collective Factorization}
Collective Factorization, described in \S\ref{sec:model}, is a general and effective method of jointly modeling all binary-valued relations in a database since it does not require custom modeling for each relation and, by learning shared universal embeddings for all entities, is able to exploit inherent dependencies between the entities for better prediction. 
Being a probabilistic model that implicitly assumes noisy observations enables the model to deal with noise in the observed relations effectively. 
This further enables the model to impute the relations as well, facilitating robustness to missing information.
We use collective factorization to improve the task of binary user preference prediction by leveraging the additional information about items and users.
Specifically, collectively modeling additional relations enables our model to learn better user and item embeddings from its attributes, category, and text reviews, facilitating more accurate user preference prediction in conjunction with past rating history. 

Learning universal embeddings for entities further enables us to predict user preferences for new items that do not contain any rating history. The embeddings for such items are learned from their category and attribute data, while user preferences for them are learned from past user ratings of similar items.  
Joint modeling of relations also enables us to impute missing entries for the item and user relations, and further, learning shared embeddings in the same latent space enables us to estimate similarity between entities that do not explicitly participate in a relation, such as similarity between review words and product categories. 

We show the efficacy of our model in the tasks described above by evaluating our model on databases from two domains, Yelp and Amazon. While both databases primarily contain user ratings and reviews, are very different from each other in the type of reviews and services they provide. Yelp contains user reviews about local businesses ranging from Restaurants to Cardiologists while Amazon is a retailer that sells products and contains user reviews about the products it sells. Apart from user ratings and reviews, both the databases contain additional information about businesses, products and users, but contain different number of entities and relations. Details about the two databases and how we convert the available information into binary-valued relations for collective factorization is described in the sections below.

\subsection{Yelp Database}
\label{sec:yelp-details}
Yelp contains rich relational data about businesses and users in the form of business attributes, categories, and user reviews and ratings. Hence, the entities present in the Yelp database we create are \emph{businesses}, \emph{categories}, \emph{attributes}, \emph{users} and \emph{review words} used by users. We denote the set of these entities by $\entB$, $\entC$, $\entA$, $\entU$ and $\entW$ respectively. Each entity in the database is represented by a $k$-dimensional embedding, as shown in the top part of Figure~\ref{cmf-yelp}. 

Each business in Yelp is categorized into a set of nearly $700$ category types according to the nature of the business. The categories available in Yelp include broad-level classes such as \emph{Doctors}, \emph{Education}, and \emph{Restaurant}, but also fine-grained descriptions such \emph{Italian}, \emph{Hookah Bars}, and \emph{Orthodontists}. 
The business category data can be viewed as a binary relation between businesses $(\entB)$ and categories $(\entC)$, and is represented as matrix \relC.

Apart from categorization, Yelp also describes various attributes for each business. Such attributes include \textit{type of parking}, \textit{delivers} (or not), \textit{noise level} and so on. We represent this relation between businesses $(\entB)$ and attributes $(\entA)$ as a binary matrix denoted by $\relA$. We transform attributes that are multi-valued into multiple binary valued attributes, for example the attribute ``\textit{Smoking}'' in the dataset has ``\textit{Yes}'', ``\textit{No}'' and ``\textit{Outdoor}'' as possible values. To represent this with binary values, we unwrap it into three separate attributes, namely ``\textit{Smoking(Yes)}'', ``\textit{Smoking(No)}'' and ``\textit{Smoking(Outdoor)}'', each of which is expressed as a binary value. 

A complex relationship between users and businesses exists in the form of ratings and text reviews. 
We represent this user-business relation in various forms. 
The ratings given by users on a $5$-scale are converted to a binary-valued preference relation between businesses $(\entB)$ and users $(\entU)$ with high ratings ($4$ and $5$) as \textit{true(1)} and low ratings ($3$ and below) as \textit{false(0)}. We denote the binary matrix representing this user preference relation by $\relR$. The relation $\relR$ thus represents the likes and dislikes of users towards businesses.
The relationship between businesses $(\entB)$ and words $(\entW)$ that are used in its reviews is represented by the relational matrix $\relBW$ in which a \textit{true(1)} value for a (business, word) tuple denotes the usage of the word for the business in at least one review. Similarly, the relation between users $(\entU)$ and the words $(\entW)$ used by them in reviews is represented as a binary matrix denoted by $\relUW$. 

Figure~\ref{cmf-yelp} gives an overview of the various relations and entities present in the Yelp database we create. It shows how different entities participate in multiple relations, which leads to their embeddings being shared among different relations. For example, the embeddings for \emph{businesses} ($\entB$) participate in modeling relations $\relC$, $\relA$, $\relR$ and $\relBW$. 

\subsection{Amazon Database}
\label{sec:amazon-details}
The Amazon database contains user ratings and review data for products sold on Amazon. Each product is categorized into multiple broad-level, intermediate, and fine-grained categories. Hence, the entities present in the Amazon database are \emph{products}, \emph{categories}, \emph{users} and \emph{review words}, denoted by \entP, \entC, \entU and \entW, respectively. Similar to Yelp, each entity in the database is represented by a $k$-dimensional embedding. 

We represent the relationship between products ($\entP$) and categories ($\entC$) as a binary-valued relational matrix, \relC. 
User ratings on Amazon are given on a $5$-scale and is converted to a binary-valued relation between products ($\entP$) and users ($\entU$) in a manner similar to Yelp. The relationship between users and products in the form of text reviews is represented as two binary-valued relations, $\relPW$ that captures the relationship between products ($\entP$) and review words ($\entW$) used for them, and $\relUW$ that represents the relation between users ($\entU$) and review words ($\entW$) used by them.

Similar to Yelp, the embeddings for different entities in the Amazon database collectively model the relations they participate in. For example, the embeddings for \emph{products} (\entP) participate in modeling relations \relR, \relC and \relPW in Amazon. The collective factorization for the Amazon database looks similar to as shown in Figure~\ref{cmf-yelp}, except that Amazon, instead of businesses, has data about products and does not contain the attribute relation. 

\begin{table}[tb]
\footnotesize
\begin{center}
\begin{tabular}{l r r r r r} 
\toprule
& $\mathbf{|\entA|}$ & $\mathbf{|\entC|}$ & $\mathbf{|\entB|}$ & $\mathbf{|\entW|}$ & $\mathbf{|\entU|}$\\
\midrule
Phoenix   & 92 & 472 & 22\,180 & 25\,277 & 102\,576 \\
Las Vegas & 92 & 416 & 14\,583 & 28\,551 & 147\,774 \\
Madison   & 77 & 176 &  2\,118 &  6\,811 &   9\,737 \\
Edinburgh & 74 & 160 &  2\,840 &  6\,830 &   2\,484 \\
\bottomrule         
\end{tabular}
\vskip 2mm
\tabcolsep=0.1cm
\footnotesize
\begin{tabular}{l r r r r r} 
\toprule
& $\mathbf{|\relA|}$ & $\mathbf{|\relC|}$ & $\mathbf{|\relR|}$ & $\mathbf{|\relBW|}$ & $\mathbf{|\relUW|}$\\
\midrule
Phoenix   & 354\,068 & 10\,468\,960 & 475\,116 & 8\,533\,231 & 12\,339\,706 \\
Las Vegas & 235\,735 &  6\,066\,528 & 556\,326 & 7\,246\,237 & 16\,598\,396 \\
Madison   &  41\,105 &     372\,768 &  35\,661 &    706\,026 &     987\,735 \\
Edinburgh &  41\,218 &     454\,400 &  20\,306 &    730\,871 &     435\,801 \\
\bottomrule         
\end{tabular}
\vskip -2mm
\caption{\label{matrix-sizes-yelp}Number of entities and observed tuples for Yelp}
\end{center}
\end{table}

\begin{table}[tb]
\tabcolsep=0.1cm
\footnotesize
\begin{center}
\begin{tabular}{l r r r} 
\toprule
& \multicolumn{1}{c}{\textbf{Arts}}  & \textbf{Electronics}    & \multirow{2}{*}{Average}   \\
& (Smallest)  & (Largest)    &    \\
\midrule
$\mathbf{|\entP|}$  & 4\,211  & 82\,067  & 33\,494 \\
$\mathbf{|\entC|}$  & 238  & 886  & 525  \\
$\mathbf{|\entU|}$  & 24\,059  & 811\,034  & 194\,078  \\
$\mathbf{|\entW|}$  & 3\,916  & 32\,086  & 12\,169  \\
\midrule
$\mathbf{|\relR|}$  & 27\,751  & 1\,196\,547  & 312\,851  \\
$\mathbf{|\relC|}$  & 20\,384  & 381\,796  & 145\,052  \\
$\mathbf{|\relPW|}$ & 382\,397  & 17\,150\,984  & 4\,424\,924  \\
$\mathbf{|\relUW|}$ & 667\,832  & 36\,621\,689  & 7\,415\,377  \\

\bottomrule         
\end{tabular}
\end{center}
\vskip -4mm
\caption{\label{matrix-sizes-amazon}Number of entities and observed tuples for Amazon}
\end{table}

\section{Experiment Setup}
\label{sec:setup}

In this section, we describe in detail the Yelp and Amazon datasets, data pre-processing and the models and experiment methodology.

\para{Datasets:} Yelp provides data from five cities namely, \textit{Phoenix}, \textit{Las Vegas}, \textit{Madison}, \textit{Edinburgh} and \textit{Waterloo}, but we focus on the first four datasets due to the small size of the \emph{Waterloo} dataset. Each of the datasets in Yelp follows the same schema, allowing evaluation of our model. For each of the datasets in Yelp we create the relational matrices, $\relR$, $\relC$, $\relA$, $\relBW$ and $\relUW$. Table~\ref{matrix-sizes-yelp} shows the number of entities participating in each relation of the various Yelp datasets, as well as the number of observed entries for each relation. 

Amazon provides datasets from $25$ broad categories of products of which we use data from $22$ of those datasets and omit \textit{Books}, \textit{Music} and \textit{Movies $\&$ TV} datasets due to their size. Like Yelp, each of the datasets in Amazon follows the same schema. For the datasets in Amazon we create $\relR$, $\relC$, $\relPW$ and $\relUW$ relation matrices. Table~\ref{matrix-sizes-amazon} shows the range of the number of entities and observed tuples per relation over the Amazon datasets.

\para{Data Pre-processing:} 
To create the $\relBW$, \relPW and \relUW matrices for Yelp and Amazon, we tokenize the reviews, remove the punctuations, numbers, and stop words, and stem the words using \citet{porter}. For evaluation purposes, we only consider words that appear in at least $10$ reviews. 
Since $\relBW$ and $\relUW$ matrices only contain observed words (all positives), we sample negative data entries in each epoch by randomly selecting a set of words that were not observed to be true for the business/user. 
The number of negative samples chosen for each relation is same as the number of observed entries for the relation. 
We found the categories \relC matrix in Yelp to be fairly comprehensive, and thus explicitly treat all unobserved entries to be negative (thus effectively \relC in Yelp is fully-observed and complete). 
Upon manual investigation, we find that the categories assigned to the products in Amazon are not always comprehensive and thus we do not treat \relC to be fully observed. Since \relC in Amazon contains only observed categories (all positives), we sample negative data entries in a manner similar to as we do for \relBW, \relPW and \relUW.
For our experiments, we only consider categories that are associated with at least $5$ businesses or products.

\para{Methods:} 
The primary benchmarks for evaluating our models will be on predicting user preferences towards businesses on Yelp and products on Amazon, in particular to study how incorporating additional information into the factorization model provides significant improvement in predictions. 
The baseline model that performs standard matrix factorization of $\relR$ independent of other relations is denoted by \textbf{R}. 
We evaluate the effect of integration of different relations on user preference prediction by factorizing combinations of different matrices collectively with the \relR matrix. An example of the model that predicts user preferences for businesses on Yelp by incorporating business categories is denoted by \textbf{R + C}.
To predict whether a relation holds between entities, we use the default logistic threshold of $0.5$ on the model probability. 
We measure the performance of our relation prediction in terms of the \emph{F1 score} defined as the harmonic mean of the precision and recall, which is a much more accurate measure than accuracy for imbalanced label distributions. 
To present a combined score for all the datasets, each in Yelp and Amazon, we aggregate all the predictions of the datasets, and compute a single F1 score over them in the micro-averaged fashion. 
Such micro-averaged F1 score would give more weightage to the larger datasets as compared to the smaller ones. One way to weigh all datasets equally is averge the F1 scores across datasets.
The value of the regularization constant, $\lambda = 0.001$, learning rate, $\gamma = 0.01$ and latent-factor dimensions $k = 30$ for Yelp and $k=5$ for Amazon is used, based on the performance on validation data.

\begin{table}[tb]
\tabcolsep=1mm
\footnotesize
\begin{center}
\begin{tabular}{l ccccccc}
\toprule
\multirow{1}{*}{}   & 
\multirow{2}{*}{{\textbf{R}}} & 
\multirow{2}{*}{{\textbf{R+C}}}   & 
\multirow{2}{*}{{\textbf{R+PW}}}  & 
\multirow{2}{*}{{\textbf{R+UW}}}  & 
\multicolumn{1}{c}{{\textbf{R+C}}}    & 
\multicolumn{1}{c}{{\textbf{R+C}}}    \\
 & 
 & 
 & 
 & 
 & 
\textbf{+PW}    & 
\textbf{+UW}    \\
\midrule
\textbf{Arts}
& 65.1
& 68.5
& 68.9
& 84.0
& 68.1
& \highest{86.4}
\\ 
\textbf{Automotive}
& 66.8
& 73.9
& 73.5
& 84.4
& 74.0
& \highest{85.7}
\\ 
\textbf{Baby}
& 67.1
& 70.7
& 71.3
& \highest{86.3}
& 70.5
& 85.8
\\ 
\textbf{Beauty}
& 71.7
& 76.0
& 75.5
& 87.1
& 75.6
& \highest{88.7}
\\ 
\textbf{Cell Phones \& Acc.}
& 55.8
& 58.9
& 58.8
& \highest{75.2}
& 58.4
& 75.1
\\ 
\textbf{Clothing Access.}
& 92.3
& 93.8
& 94.0
& 93.6
& 93.9
& \highest{96.2}
\\ 
\textbf{Electronics}
& 66.5
& 71.0
& 70.8
& 82.4
& 70.8
& \highest{84.2}
\\ 
\textbf{Gourmet Foods}
& 64.5
& 72.1
& 71.9
& 85.2
& 72.2
& \highest{88.3}
\\ 
\textbf{Health}
& 67.0
& 71.8
& 71.7
& 84.5
& 71.6
& \highest{86.4}
\\ 
\textbf{Home Kitchen}
& 68.1
& 72.6
& 72.7
& 83.9
& 72.9
& \highest{86.4}
\\ 
\textbf{Industrial Scientific}
& 91.1
& 93.4
& 93.7
& 94.5
& 93.7
& \highest{96.7}
\\ 
\textbf{Jewelry}
& 68.0
& 74.8
& 74.3
& 81.5
& 74.5
& \highest{87.4}
\\ 
\textbf{Musical Instruments}
& 61.1
& 71.1
& 70.9
& 85.3
& 71.0
& \highest{88.0}
\\ 
\textbf{Office Products}
& 63.9
& 67.3
& 67.2
& 83.1
& 67.5
& \highest{85.1}
\\ 
\textbf{Pet Supplies}
& 65.1
& 70.2
& 69.7
& 84.2
& 70.1
& \highest{85.9}
\\ 
\textbf{Shoes}
& 97.6
& 96.0
& 95.9
& 95.0
& 95.8
& \highest{97.6}
\\ 
\textbf{Software}
& 54.7
& 58.3
& 58.4
& 68.0
& 58.3
& \highest{71.7}
\\ 
\textbf{Sports Outdoors}
& 71.9
& 76.6
& 76.9
& 85.5
& 76.4
& \highest{88.7}
\\ 
\textbf{Tools \& Home Imp.}
& 66.5
& 72.3
& 72.4
& 84.1
& 72.7
& \highest{86.6}
\\ 
\textbf{Toys Games}
& 65.8
& 70.7
& 70.6
& 84.3
& 70.5
& \highest{86.8}
\\ 
\textbf{Video Games}
& 68.1
& 71.1
& 70.8
& 82.6
& 71.2
& \highest{83.7}
\\ 
\textbf{Watches}
& 61.6
& 64.2
& 63.7
& 85.6
& 64.1
& \highest{87.7}
\\ 
\midrule
\textbf{Combined}
& 72.1
& 75.0
& 75.8
& 85.5
& 75.1
& \highest{87.3}
\\ 
\bottomrule         
\end{tabular}
\end{center}
\vskip -5mm
\caption{\label{rating-held-out-amazon-nobias} \textbf{Held-out User Preference prediction on Amazon:} F1 of predicting held-out user preferences from \relR. The models being evaluated vary in the number of relations modeled when learning the factors, with additional relations often resulting in more accurate models across datasets.}
\end{table}

\begin{table}[tb]
\tabcolsep=1mm
\footnotesize
\begin{center}
\begin{tabular}{l ccccc}
\toprule
 & 
Phnx. & 
L.Vegas &
Madison &
Ednbgh.  & 
\textbf{Combined} \\ 
\midrule
\textbf{R}
& 72.2
& 70.5
& 68.5
& 75.2
& 71.3
\\ \addlinespace[1mm]
\textbf{R+A}
& 73.7
& 71.1
& 70.5
& 76.3
& 72.3
\\
\textbf{R+BW}
& 74.2
& 70.8
& 71.4
& 75.8
& 72.4
\\ 
\textbf{R+C}
& 75.1
& 72.2
& 72.2
& 76.4
& 73.6
\\ 
\textbf{R+UW}
& 80.0
& 78.5
& 78.5
& 78.5
& 79.2
\\ 
\addlinespace[1mm]
\textbf{R+A+C}
& 74.2
& 71.1
& 70.7
& 74.8
& 72.5
\\ 
\textbf{R+A+BW}
& 74.1
& 70.7
& 70.8
& 76.5
& 72.3
\\ 
\textbf{R+C+BW}
& 74.3
& 70.9
& 71.7
& 77.2
& 72.6
\\ 
\textbf{R+A+UW}
& 80.0
& 78.5
& 78.9
& 79.1
& 79.2
\\
\textbf{R+C+UW}
& \highest{80.9}
& \highest{79.9}
& \highest{80.8}
& \highest{79.0}
& \highest{80.4}
\\ 
\addlinespace[1mm]
\textbf{R+A+C+BW}
& 73.9
& 70.8
& 71.1
& 76.4
& 72.3
\\ 
\textbf{R+A+C+UW}
& 80.8
& 79.3
& 77.9
& 78.5
& 79.9
\\ 
\bottomrule         
\end{tabular}
\end{center}
\vskip -4mm
\caption{\label{rating-held-out} \textbf{Held-out User Preference prediction on Yelp:} F1 on the predicting held-out ratings from \relR for Yelp as the amount and type of additional information is varied.}
\end{table}

\begin{figure}[tb]
\centering
    \begin{subfigure}[tb]{0.35\textwidth}
        \centering
        \includegraphics[width=\columnwidth,trim=0mm 1mm 3mm 6mm,clip]{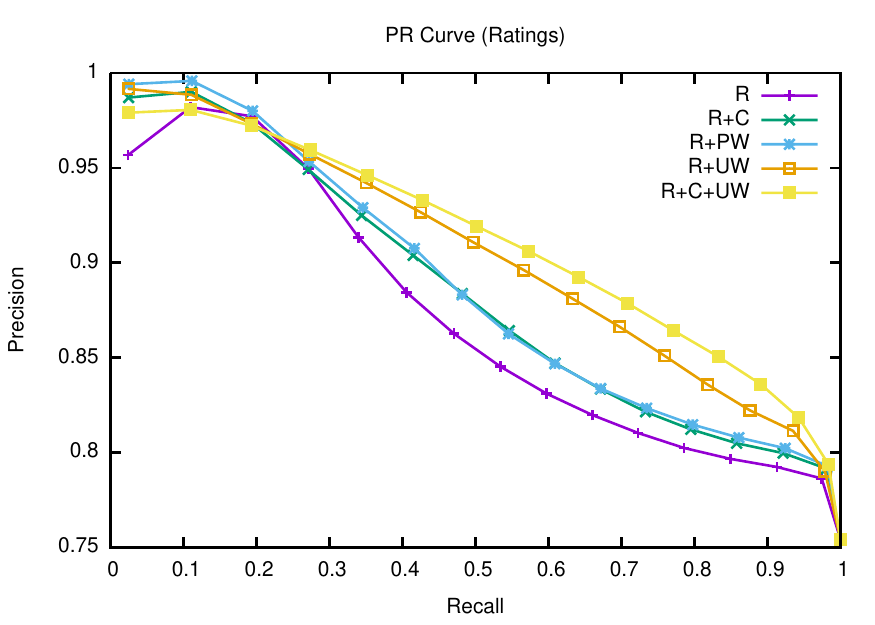}
        \caption{Amazon}
        \label{fig:pr-amazon-ho}
    \end{subfigure}
    \qquad
    \begin{subfigure}[tb]{0.35\textwidth}
        \centering
        \includegraphics[width=\columnwidth]{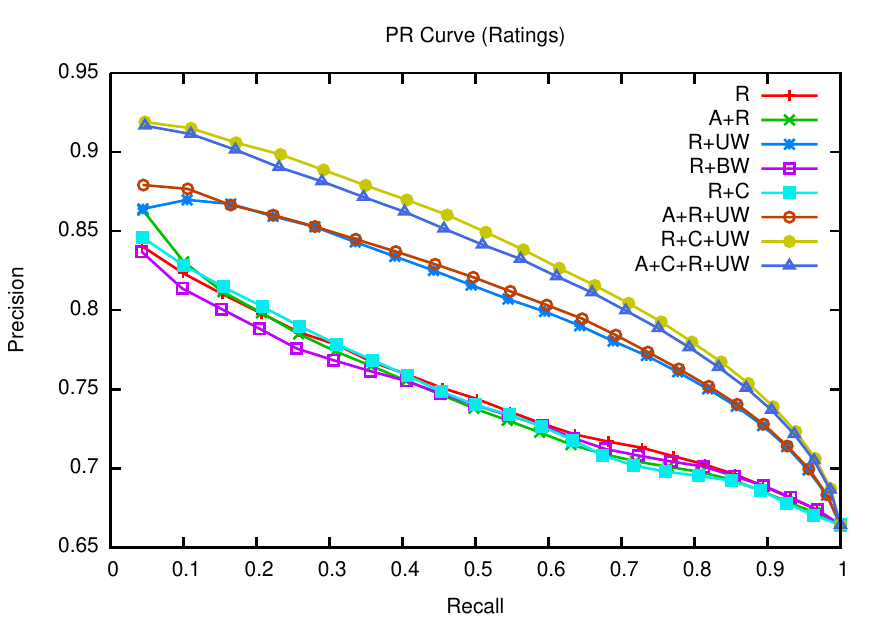}
        \caption{Yelp}
        \label{fig:pr-yelp-ho}
    \end{subfigure}
    \caption{Precision/Recall Curves for Held Out User Preferences}
    \label{fig:pr:held}
\end{figure}

\section{Results}
\label{sec:results}
In this section, we evaluate the effect of incorporating relational information in predicting user preferences.
First, we present the accuracy of predicting preferences on \emph{held-out} items in \S\ref{sec:results:held-out} to test the performance when the entities already contain a few observed ratings. 
We further investigate the performance of different models on \emph{cold-start} estimation in \S\ref{sec:results:cold-start}, where, for example, we predict user preferences for businesses with \emph{no} past ratings or reviews available. 
One of the major advantages of modeling all relations jointly is the ability to complete other missing relational matrices while at the same time predicting user preferences. We investigate the performance of our collective factorization model in predicting \emph{business attributes} in Yelp and \emph{product categories} in Amazon in \S\ref{sec:results:imputing-db} by utilizing category, ratings, and reviews.
Finally, utilizing the fact that the model embeds all entity types in the same $k$-dimensional space, we present visualizations in \S\ref{sec:results:qualitative} that explore similarities between entities for which explicit relations are not observed. 

\subsection{User Preference Prediction}
\label{sec:results:held-out}
The most important problem in recommendation systems is to be able to predict user preferences for products/businesses. 
To show how our model improves significantly on predicting user preferences by leveraging additional information for existing entities, we carry out evaluation of held-out predictions. 
To split the Yelp and Amazon data into training, validation and test sets for evaluation, we randomly choose $70\%$ and $80\%$ of the observed ratings from Yelp and Amazon respectively, for training and equally divide the remaining data into validation and test sets. We present the performance of different models by varying the user relations, business relations for Yelp and product relations for Amazon available during training for user preference prediction. 

\para{User Preference Prediction on Amazon:}
We expect that incorporation of additional information about \emph{products} and \emph{users} on Amazon, such as categories and text reviews should improve the prediction of user preferences. 
Results for collective factorization of combinations of various relational matrices containing information about products and users on Amazon, with the \relR matrix are shown in Table~\ref{rating-held-out-amazon-nobias}.
Our baseline model achieves an \emph{F1 score} of $72.1\%$ on predicting user preferences and gains of $4\%$ and $5.1\%$ are observed when incorporating information about the products in terms of its categories $(\relC)$ and review words used for them $(\relPW)$, respectively. 
Significant improvement of $18.5\%$ from the baseline is obtained by incorporating the relationship between the users and the review words they use $(\relUW)$. It is clear from this, that the user reviews are quite indicative of their likes and dislikes for various aspects of a product, which further helps to predict user preferences. 
Additional increase of $2.1\%$ is obtained by additionally incorporating \emph{product category} $(\relC)$ information on top of \emph{user-words} $(\relUW)$ relation. Figure~\ref{fig:pr-amazon-ho} shows that, even though different models perform similarly in the high precision and recall regions, models that incorporate more information dominate the majority of the plot. 

\para{User Preference Prediction on Yelp:}
Table~\ref{rating-held-out} shows the accuracy of our collective factorization model in user preference prediction on Yelp by combining various relational matrices about businesses and users with the $\relR$ matrix.  
Our baseline model achieves an \emph{F1 score} of $71.3\%$, with an increase of $1.4\%$ when incorporating information about the businesses in terms of its attributes $(\relA)$ or review words used for them $(\relBW)$. 
Incorporating business categories $(\relC)$ improves upon the baseline model by $3.22\%$. 
Similar to Amazon, significant improvement of $11.07\%$ from the baseline is obtained by incorporating relationship between the users and their review words $(\relUW)$. 
Further increase of $1.5\%$ is obtained by incorporating \emph{business category} $(\relC)$ information on top of \emph{user-words} $(\relUW)$ relation. 
When adding information about \emph{business attributes} along with \emph{business categories} and \emph{user-words} relations, we find that the prediction accuracy falls only slightly by $0.62\%$. 
A reason for this may be a lack of dependence between user preferences and business attributes, and thus modeling attributes along with preferences slightly affects the accuracy. 
The precision/recall curves in Figure~\ref{fig:pr-yelp-ho} show how incorporating different kinds of information about users and businesses affect user preference prediction. 
It is clear that \emph{user-word} relation provides higher gains than incorporating information about businesses, but more importantly, integrating information about both businesses and users achieves the best performance.


As in clear from the above results, our collective factorization model is efficient in integrating additional information about products, businesses and users when predicting user preferences for users with available rating history. 
In both the databases it is seen that incorporating the \emph{product-word} $(\relPW)$ or \emph{business-word} $(\relBW)$ relation helps improve user preference prediction that shows, incorporation of words used for products and businesses helps learn better embeddings for both. 
Additionally, the increase by incorporation of the \emph{category} information $(\relC)$, suggests that the addition of categories helps the model learn user biases towards certain categories along with their other preferences. 
Significant improvements in user preference prediction by integration of knowledge about the words used by users $(\relUW)$ shows the importance of reviews written by users in learning better embeddings for users by learning learning in-depth likes and dislikes of users that are not reflected in just the ratings but in the detailed reviews written.

Including the per-entity biases and a global offset for each relational matrix gave similar accuracy on user-preference prediction when factorizing all relations collectively. The baseline for each Yelp and Amazon was higher than the current baseline and thus the gains from additional information were reduced. 

\begin{table}[tb]
\tabcolsep=1mm
\footnotesize
\begin{center}
\begin{tabular}{l@{\hskip 4mm} c@{\hskip 4mm} c@{\hskip 4mm} c@{\hskip 4mm} c}
\toprule
\multirow{2}{*}{} & 
\textbf{R*}         & 
\textbf{R+UW}         & 
\textbf{R+C}         & 
\textbf{R+C+UW}    \\
\midrule
\textbf{Arts}
& 60.8
& 64.5
& 65.4
& \highest{86.9}
\\ 
\textbf{Automotive}
& 58.7
& 60.0
& 73.0
& \highest{87.7}
\\ 
\textbf{Baby}
& 60.8
& 57.9
& 70.7
& \highest{85.6}
\\ 
\textbf{Beauty}
& 60.7
& 56.9
& 75.9
& \highest{86.4}
\\ 
\textbf{Cell Phones \& Acc.}
& 54.1
& 53.0
& 56.8
& \highest{71.3}
\\ 
\textbf{Clothing Accessories}
& 59.4
& 62.3
& 94.1
& \highest{95.0}
\\ 
\textbf{Electronics}
& 59.3
& 57.7
& 71.3
& \highest{83.8}
\\ 
\textbf{Gourmet Foods}
& 61.7
& 56.0
& 71.6
& \highest{88.7}
\\ 
\textbf{Health}
& 59.4
& 61.2
& 71.0
& \highest{86.3}
\\ 
\textbf{Home Kitchen}
& 58.7
& 60.6
& 72.4
& \highest{84.2}
\\ 
\textbf{Industrial Scientific}
& 61.4
& 62.0
& 91.9
& \highest{96.0}
\\ 
\textbf{Jewelry}
& 61.5
& 64.9
& 76.7
& \highest{77.3}
\\ 
\textbf{Musical Instruments}
& 59.8
& 60.3
& 70.5
& \highest{88.3}
\\ 
\textbf{Office Products}
& 59.8
& 58.7
& 67.2
& \highest{83.2}
\\ 
\textbf{Pet Supplies}
& 59.1
& 59.8
& 70.9
& \highest{85.1}
\\ 
\textbf{Shoes}
& 59.2
& 58.7
& 95.8
& \highest{97.1}
\\ 
\textbf{Software}
& 51.6
& 54.2
& 59.0
& \highest{72.6}
\\ 
\textbf{Sports Outdoors}
& 61.8
& 62.1
& 77.6
& \highest{87.6}
\\ 
\textbf{Tools \& Home Impr.}
& 59.3
& 61.9
& 72.5
& \highest{85.8}
\\ 
\textbf{Toys Games}
& 61.3
& 59.7
& 70.5
& \highest{84.8}
\\ 
\textbf{Video Games}
& 59.3
& 61.3
& 67.9
& \highest{77.7}
\\ 
\textbf{Watches}
& 60.6
& 59.6
& 65.6
& \highest{85.9}
\\ 
\midrule
\textbf{Combined}
& 59.6
& 59.9
& 75.7
& \highest{86.7}
\\ 
\bottomrule         
\end{tabular}
\end{center}
\vskip -4mm
\caption{\label{rating-cold-start-amazon-nobias} \textbf{Cold-Start User Preference prediction on Amazon:} F1 scores of different collective factorization models in predicting the user preferences for products on Amazon for which \emph{no} ratings or reviews are observed. *Conventional matrix factorization, \textbf{R}, is a trivial straw-man in that it does not have any way to differentiate amongst cold-start businesses.}
\end{table}

\subsection{Cold-Start Evaluation}
\label{sec:results:cold-start}
One of the major challenges faced by recommendation systems is to predict ratings for new products, businesses and users for which no reviews or ratings have been observed. This problem is not just specific to recommendation systems, but common to all relation prediction frameworks. Most of the factorization models for relation prediction fail to incorporate information about entities from relations, apart from the relation to be predicted, and thus provide poor cold-start performance. 

Our collective factorization model benefits greatly from learning shared factors for entities by leveraging all sources of information about the entity. 
Hence, in the absence of observed data for a particular relation, embeddings learned from other relations can still be used to predict the relation. 
Specifically, we show that our model can learn factors for products and businesses for which no reviews or ratings were observed from its categories and attributes, and use them to predict user preferences. For evaluation on Amazon, we withhold \emph{all} observed cells of the relation \relR for a random $10\%$ of the products. We use $90\%$ of the data for the remaining products for training and the rest for validation. 
We split the Yelp data similarly, by withholding \emph{all} observed cells of the relation \relR for a random $10\%$ of the businesses. Apart from the variety of collective models, we also include the uninformative straw man that has the same prediction for all cold-start products on Amazon and businesses on Yelp in user preference prediction, evaluated by computing the \emph{F1 score} when the the factors for new products and businesses are randomly initialized to small values. 

\begin{figure}[tb]
\centering
    \begin{subfigure}[tb]{0.35\textwidth}
        \centering
        \includegraphics[width=\columnwidth,trim=0mm 1mm 3mm 6mm,clip]{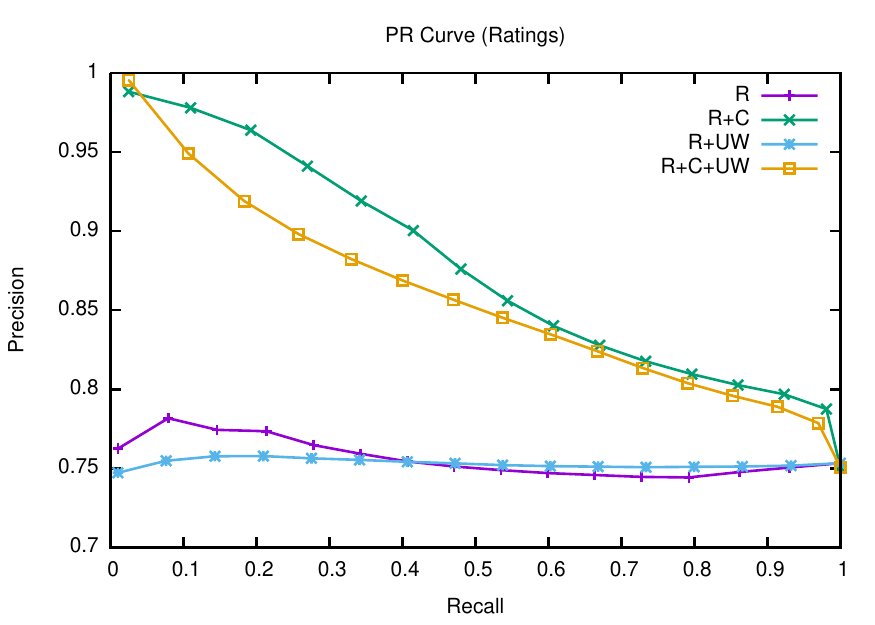}
        \caption{Amazon}
        \label{fig:pr-amazon-cs}
    \end{subfigure}
    \\
    \begin{subfigure}[tb]{0.35\textwidth}
        \centering
        \includegraphics[width=\columnwidth,trim=0mm 1mm 3mm 6mm,clip]{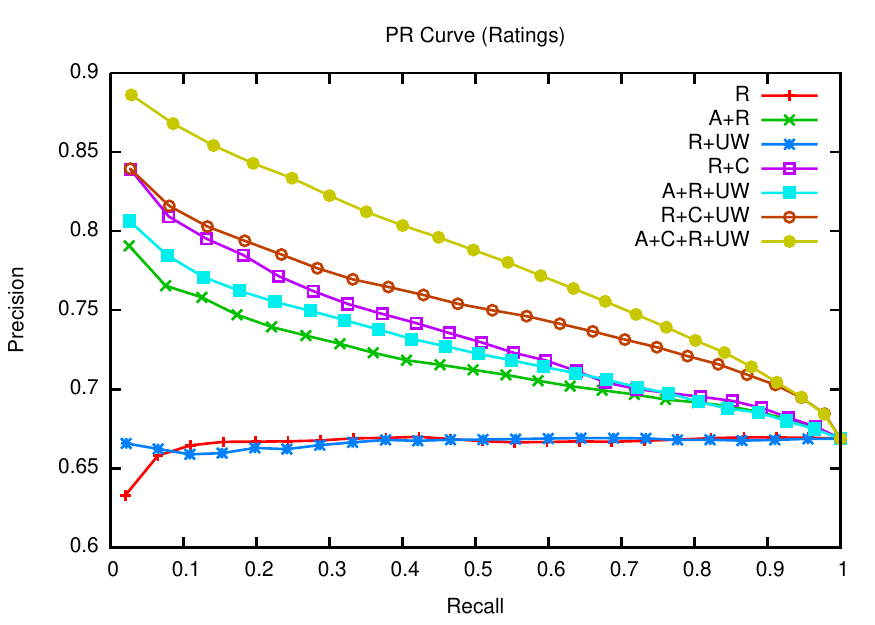}
        \caption{Yelp}
        \label{fig:pr-yelp-cs}
    \end{subfigure}
    \caption{Precision/Recall Curves for Cold-Start User Preferences}
    \label{fig:prcs}
\end{figure}

\para{Product Cold-Start on Amazon:}
Predicting user preferences for new products \emph{a priori} is an exciting problem since it can help sellers on Amazon quickly identify the target audiences that would like the products. Users that have biases towards certain categories prefer products that cater to their needs. Using our collective factorization model, we can integrate category knowledge for new products, along with user reviews and preferences for existing products, to predict preferences for new ones. 
Table~\ref{rating-cold-start-amazon-nobias} shows the performance of different models that vary in information being used to predict user preferences for products that do not contain any rating history in the database. The results corroborate the fact that learning good embeddings just for users (via $\relUW$) is not enough to predict user preferences. 
We find that incorporating product information in terms of its categories $(\relC)$ obtains an accuracy as high as $75.7\%$, and further, integrating of user-word relation $(\relUW)$ results in prediction F1 as high as $86.7\%$. 
On the other hand, Figure~\ref{fig:pr-amazon-cs} suggests that adding categories by itself would obtain a higher accuracy than adding both category and user-word relations if the threshold is tuned.

\para{Business Cold-Start on Yelp:}
Table~\ref{rate-bus-cold} shows how collective factorization is able to use \emph{business category} $(\relC)$ and \emph{business attribute} $(\relA)$ relation to learn good embeddings for businesses and predict user preferences for businesses with no available rating history. As expected, learning good embeddings just for users (via \relUW) is not enough to predict user preferences. 
Incorporation of information about businesses in terms of attributes $(\relA)$ and categories $(\relC)$ obtains prediction F1 as high as $71.1\%$ and $73.5\%$ respectively. 
Collectively integrating all information about businesses and users simultaneously, such as categories $(\relC)$, attributes $(\relA)$ and user-word relation $(\relUW)$ leads to prediction F1 as high as $78.2\%$. Figure\ref{fig:pr-yelp-cs} shows how integrating additional information about both businesses and users outperforms models with relatively less information.

By obtaining results close to those in \S~\ref{sec:results:held-out}, we demonstrate that the collective factorization model is able to almost completely overcome the lack of existing user preferences for products and businesses by utilizing other relations.

\begin{table}[tb]
\tabcolsep=1mm
\footnotesize
\begin{center}
\begin{tabular}{l ccccc}
\toprule
 & 
Phnx. & 
L.Vegas &
Madison &
Ednbgh.  & 
\textbf{Combined} \\ 
\midrule
\textbf{R*}
& 57.0
& 56.9
& 58.0
& 56.3
& 57.0
\\
\addlinespace[1mm]
\textbf{R+UW}
& 57.8
& 57.0
& 56.4
& 59.7
& 57.4
\\
\textbf{R+A}
& 72.8
& 69.4
& 70.6
& 74.2
& 71.1
\\ 
\textbf{R+C}
& 75.1
& 72.1
& 72.3
& 75.5
& 73.5
\\
\addlinespace[1mm]
\textbf{R+A+C}
& 74.9
& 71.2
& 71.7
& \highest{77.6}
& 73.0
\\
\textbf{R+A+UW}
& 78.1
& 74.5
& \highest{78.4}
& 75.1
& 76.3
\\
\textbf{R+C+UW}
& 79.6
& 76.2
& 74.2
& 70.0
& 77.6
\\
\addlinespace[1mm]
\textbf{R+A+C+UW}
& \highest{79.6}
& \highest{77.4}
& 77.8
& 72.6
& \highest{78.2}
\\

\bottomrule         
\end{tabular}
\end{center}
\vskip -4mm
\caption{\label{rate-bus-cold} \textbf{Cold-Start User Preference prediction on Yelp:} F1 scores of different collective factorization models in predicting the user preferences for businesses for which \emph{no} ratings or reviews are observed. *For matrix factorization \textbf{R} all cold-start business rows are empty, and thus is a straw-man.}
\end{table}

\begin{table}[tb]
\tabcolsep=1mm
\footnotesize
\begin{center}
\begin{tabular}{l@{\hskip 4mm} c@{\hskip 4mm} c@{\hskip 4mm} c@{\hskip 4mm} c}
\toprule
\multirow{2}{*}{} & 
\textbf{C*}         & 
\textbf{C+R}         & 
\textbf{C+PW}         & 
\textbf{C+R+PW}    \\
\midrule
\textbf{Arts}
& 49.5
& 52.1
& \highest{85.4}
& 83.1
\\ 
\textbf{Automotive}
& 49.9
& 53.9
& 87.6
& \highest{87.8}
\\ 
\textbf{Baby}
& 52.3
& 54.4
& \highest{88.2}
& 88.0
\\ 
\textbf{Beauty}
& 49.7
& 58.5
& \highest{89.8}
& 89.1
\\ 
\textbf{Cell Phones \& Acc.}
& 48.6
& 52.7
& \highest{90.5}
& 90.0
\\ 
\textbf{Clothing Accessories}
& 49.8
& 67.7
& 90.1
& \highest{90.1}
\\ 
\textbf{Electronics}
& 49.3
& 57.8
& 90.4
& \highest{90.5}
\\ 
\textbf{Gourmet Foods}
& 49.5
& 57.3
& \highest{87.4}
& 87.1
\\ 
\textbf{Health}
& 49.7
& 57.8
& 89.5
& \highest{89.8}
\\ 
\textbf{Home Kitchen}
& 49.9
& 57.3
& 89.9
& \highest{89.7}
\\ 
\textbf{Industrial Scientific}
& 50.3
& 66.7
& \highest{89.7}
& 89.6
\\ 
\textbf{Jewelry}
& 49.2
& 59.0
& 83.9
& \highest{84.9}
\\ 
\textbf{Musical Instruments}
& 50.3
& 53.0
& 86.3
& \highest{87.0}
\\ 
\textbf{Office Products}
& 49.0
& 53.3
& 86.6
& \highest{86.8}
\\ 
\textbf{Pet Supplies}
& 50.4
& 59.4
& \highest{90.6}
& 90.3
\\ 
\textbf{Shoes}
& 50.0
& 77.2
& 90.3
& \highest{90.6}
\\ 
\textbf{Software}
& 49.6
& 53.7
& \highest{87.8}
& 86.6
\\ 
\textbf{Sports Outdoors}
& 49.6
& 59.6
& \highest{89.0}
& 88.8
\\ 
\textbf{Tools \& Home Impr.}
& 49.7
& 55.8
& 87.7
& \highest{87.8}
\\ 
\textbf{Toys Games}
& 49.6
& 59.8
& 89.2
& \highest{89.3}
\\ 
\textbf{Video Games}
& 49.1
& 61.4
& 89.8
& \highest{90.3}
\\ 
\textbf{Watches}
& 51.1
& 53.7
& \highest{84.6}
& 83.8
\\ \midrule
\textbf{Combined}
& 49.7
& 59.4
& \highest{88.9}
& 88.8
\\ 

\bottomrule         
\end{tabular}
\end{center}
\vskip -4mm
\caption{\label{category-cold-start-amazon-nobias} \textbf{Imputing Product Categories on Amazon:} F1 evaluation for imputing the categories for products on Amazon for which \emph{no} category data was observed. *Matrix factorization \textbf{C}, cannot differentiate at all amongst the cold-start products.}
\end{table}

\begin{table}[tb]
\tabcolsep=1mm
\footnotesize
\begin{center}
\begin{tabular}{l ccccc}
\toprule
 & 
Phnx. & 
L.Vegas &
Madison &
Ednbgh.  & 
\textbf{Combined} \\ 
\midrule
\textbf{A*}
& 38.9
& 37.6
& 38.2
& 34.3
& 38.1
\\ \addlinespace[1mm]
\textbf{A+R}
& 57.5
& 53.5
& 51.3
& 53.4
& 55.5
\\
\textbf{A+C}
& 79.6
& 77.4
& 74.0
& 76.2
& 78.3
\\ 
\textbf{A+BW}
& 81.8
& \highest{81.8}
& \highest{77.5}
& 72.7
& 81.0
\\ \addlinespace[1mm]
\textbf{A+R+C}
& 78.7
& 75.9
& 72.9
& 74.7
& 77.2
\\ 
\textbf{A+R+BW}
& 82.1
& 80.6
& 77.2
& 74.4
& 80.9
\\ 
\textbf{A+C+BW}
& 82.1
& 81.0
& 77.1
& \highest{76.6}
& \highest{81.1}
\\
\addlinespace[1mm]
\textbf{A+R+C+BW}
& \highest{82.1}
& 80.6
& 77.3
& 75.7
& 80.9
\\
\bottomrule         
\end{tabular}
\end{center}
\vskip -4mm
\caption{\label{att-bus-cold} \textbf{Imputing Business Attributes on Yelp:} Performance of different collective factorization models in predicting attributes for business without any observed attributes. *Similar to other cold-start results, matrix factorization \textbf{A} here is also an uninformative baseline that has the same prediction for all business attributes.}
\end{table}

\begin{figure}[b!]
    \centering
        \includegraphics[width=0.5\textwidth]{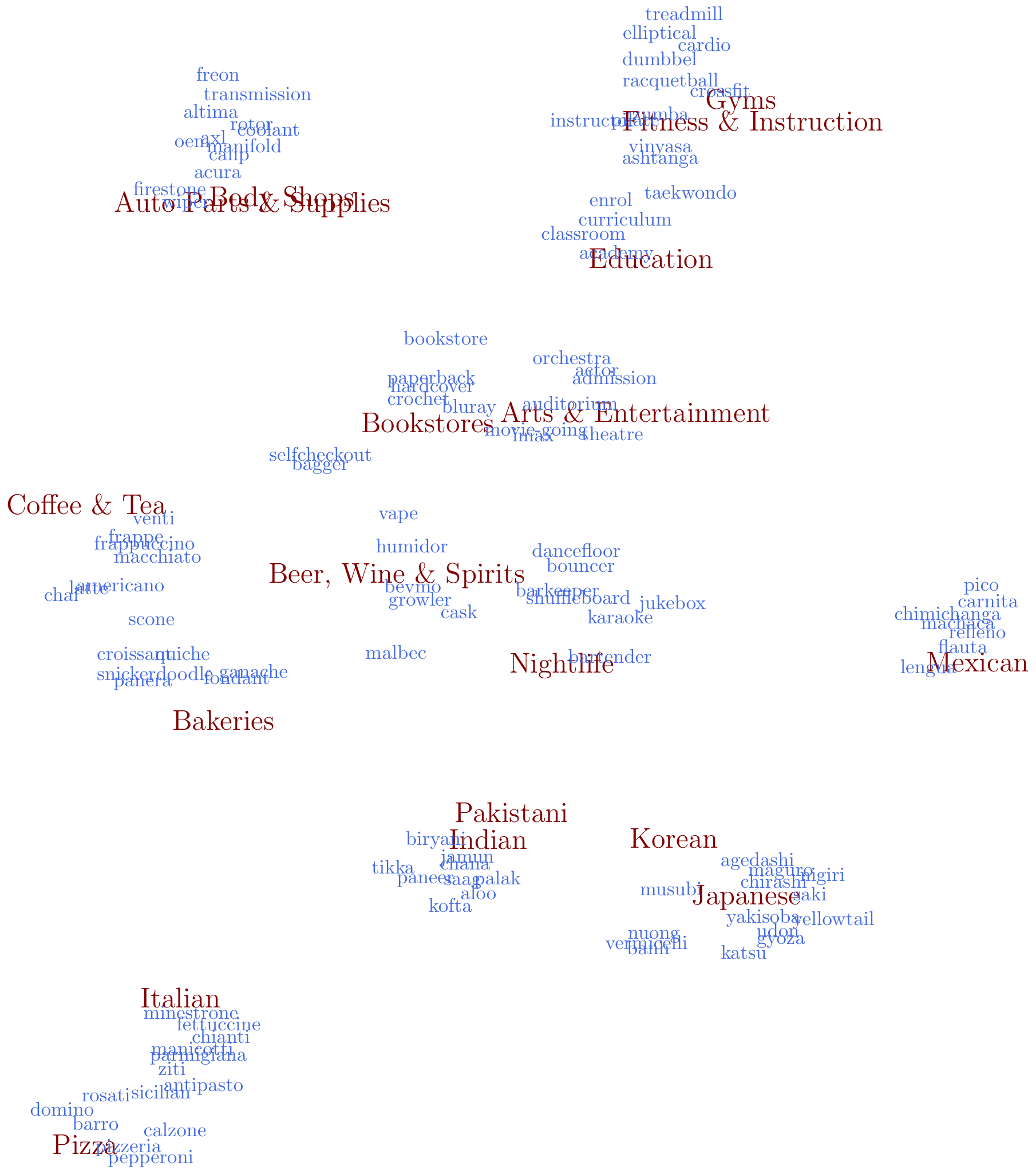}
    \caption{Visualization of \textcolor{Maroon}{Category} and \textcolor{RoyalBlue}{Review Word} factors, showing similarity of factors for semantically-similar words and categories. Best viewed in color.}
    \label{fig:tsne:cat}
\end{figure}

\subsection{Imputing Missing Entries in Database}
\label{sec:results:imputing-db}
Joint factorization of all relations in the database by learning shared universal embeddings for entities to predict user preferences, also enables us to predict missing entries in the incomplete relational matrices simultaneously while they help in predicting user preferences. For example integration of \emph{business category} $(\relC)$ and \emph{business-review words} $(\relBW)$ relation in Yelp while aiding the prediction of user preferences also helps in completing missing attributes for businesses in \relA in the Yelp database. This potential of our collective factorization model to impute missing entries in the database has several advantages. For example, predicting missing attributes in the Yelp database helps in completing the database, more accurate prediction of user preferences, and also helps users make more informed decisions when choosing between businesses.

In this section, we evaluate our model's efficacy in completion of additional relational information of entities by predicting \emph{product categories} $(\relC)$ in Amazon and \emph{business attributes} $(\relA)$ in Yelp. 
Similar to \S\ref{sec:results:cold-start}, we present \emph{cold-start} evaluations, where we show how additional information about products and businesses can be leveraged to predict categories for products in Amazon and attributes for businesses in Yelp without any observed category and attribute information, respectively. 
For evaluation, we withhold \emph{all} observed cells of the relation \relC and \relA for a random $10\%$ of the products on Amazon and businesses in Yelp, respectively. We use $90\%$ of the remaining data for training and the rest for validation. 
Apart from the variety of collective models, similar to \S\ref{sec:results:cold-start}, we also include the uninformative straw man that has the same prediction for all cold-start products on Amazon for product category prediction and businesses on Yelp for business attribute prediction. 

\para{Product Category Completion on Amazon:} 
Table~\ref{category-cold-start-amazon-nobias} shows the accuracy of our model in predicting missing product category data for products with no observed category data during training by learning useful product embeddings from other relations. 
We see that incorporating \emph{products-review words} $(\relPW)$ relation helps in predicting product categories with a F1 score of $88.9\%$. Additionally incorporating the user preference data does not improve the accuracy which suggests that it is the reviews that help the most in learning good embeddings for products when predicting categories they belong to.


\para{Business Attribute Completion on Yelp:}
Quite surprisingly, $15.91\%$ of the total $42\,151$ businesses in the Yelp database do not contain any information about their attributes. 
In Table~\ref{att-bus-cold} we show the accuracy of our model on attribute completion for businesses with no attributes observed during training. 
As expected, incorporating business-category $(\relC)$ and business-word ($\relBW$) relations helps the most in predicting attributes for new businesses. \emph{F1 score} as high as $78.3\%$ is achieved on incorporating the $\relC$ relation and further addition of $\relBW$ relation achieves a \emph{F1 score} of $81.1\%$. Integrating ratings data on top, doesn't affect the model a lot and obtains an \emph{F1 score} of $80.9\%$.

\subsection{Jointly Visualizing Entity Embeddings}
\label{sec:results:qualitative}
Finding relationships and similarity between entities that do not participate in the same relation in the database schema is a challenging problem in relation learning. Similarity between entities has important applications in data visualization and developing intuitive user interfaces, amongst others. Since our model defines embeddings for all entities over the same $k$-dimensional space, we can compute similarities between any pairs of entities even if they do not appear in the same relation. 
For example, reviews, along with indicating user preferences, also contain information about the business categories and attributes. In this section, we show how the learnt embeddings for review words, categories and attributes in Yelp that reveal similarity between them. We project the embeddings of a select subset of categories, attributes and a subset of similar review words onto a $2$-dimensional plot using the \emph{t-Distributed Stochastic Neighbor Embedding (t-SNE)}~\cite{tsne} technique for dimensionality reduction.
This is a randomized, approximate technique that attempts to maintain the distances between entities in $k$ dimensions when projecting them to two dimensions.
The vectors used here for categories, attributes and review words are obtained by collectively factorizing $\relA$, $\relR$, $\relC$ and $\relBW$ relations on the Yelp \emph{Phoenix} data.

\para{Visualizing Categories and Review Words:}
The efficacy of our model in learning inter-category similarity and similarity between categories and review words in Yelp is shown in Figure~\ref{fig:tsne:cat}. For example, our model is able to learn that \emph{Indian} and \emph{Pakistani} cuisines and \emph{Korean} and \emph{Japanese} cuisines are similar to each other, \emph{Gyms} and \emph{Fitness $\&$ Instruction} categories for businesses are similar and \emph{Beer, Wine $\&$ Spirits} is related to \emph{Nightlife}. Our model also learns the semantic similarity between categories and review words. For example, words closest to \emph{Auto Parts \& Supplies} are \emph{rotor}, \emph{coolant}, \emph{wiper}, \emph{transmission} and closest to \emph{Arts \& Entertainment} are \emph{auditorium}, \emph{theatre}, \emph{imax}, \emph{movie-going} and \emph{orchestra}. For categories related to food, our model learns the names of dishes as being closest to categories, suggesting that the users mostly talk about the dishes when reviewing restaurants. For example, the words closest to  \emph{Mexican} are \emph{carnita}, \emph{flauta}, \emph{chimichanga}, and \emph{relleno}, \emph{Coffee $\&$ Tea} are \emph{frappe}, \emph{chai}, \emph{macchiato}, and \emph{frappuccinno}, and \emph{Bakeries} are \emph{scone}, \emph{croissant}, and \emph{quiche}.
Our model is also able to learn word embeddings in such a manner that same words used in reviews for dissimilar businesses are approximately between both the categories. For example, words like \emph{enroll}, \emph{taekwondo} and \emph{curriculum}, which may belong to reviews of both education related businesses and fitness centers, lie in between the \emph{Education} and \emph{Fitness \& Instruction} category. Similar observations are made in words that are close to \emph{Bakeries} and \emph{Coffee \& Tea} categories.



\begin{figure}[tb]
    \centering
    \includegraphics[width=0.5\textwidth]{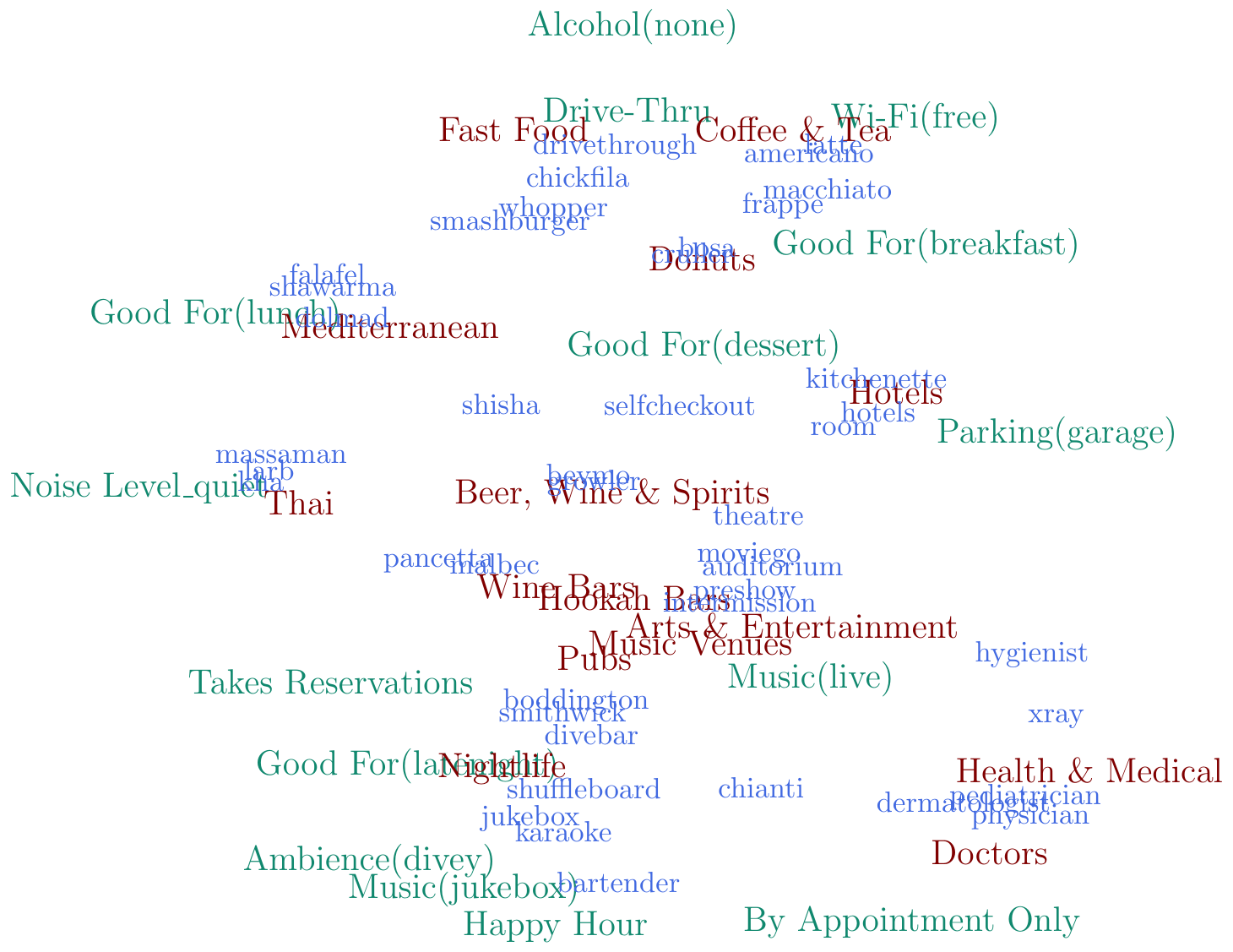}
    \caption{Visualization of \textcolor{Maroon}{Categories}, \textcolor{PineGreen}{Attributes}, \& \textcolor{RoyalBlue}{Words}. Best viewed in color.}
    \label{fig:tsne:attr-cat}
\end{figure}

\para{Visualizing Categories, Attributes, and Words:}
In Figure~\ref{fig:tsne:attr-cat}, we plot a subset of categories and the attributes and review words that are close to them. 
From the figure, we see that our model learns similar embeddings for attributes that co-exist for certain types of businesses. The proximity of \emph{Ambience(divey)} and \emph{Good For(latenight)} attributes suggests that places that have a divey ambience are good for late nights. Also, businesses that have a happy hour mostly also have a Jukebox is shown by the proximity of \emph{Happy Hour} and \emph{Music (Jukebox)}. 
The proximity of the categories \emph{Fast Food} and \emph{Coffee \& Tea} to the attributes \emph{Drive-Thru}, \emph{Wi-fi(free)} and \emph{Alcohol(none)}, category \emph{Doctors} to attribute \emph{By Appointment Only}, and the category \emph{Arts \& Entertainment} to attribute \emph{Music(live)}, all demonstrate that the model is able to learn how certain categories of businesses are most likely to have certain attributes. 
We also see from the figure that the even though the reviews do not explicitly talk about the attributes and categories, our model is able to capture the similarity between them simultaneously. For example, words like \emph{jukebox}, \emph{karaoke}, \emph{bartender}, \emph{chianti} lie close to attributes \emph{Good For(latenight)}, \emph{Happy Hour} and the category \emph{Nightlife}.  

\section{Related Work}
\label{sec:related}
This work builds upon a large and growing area of machine learning applied to recommendation systems and modeling of structured datasets.
We describe a subset of these approaches that are directly related to, and inspired, our proposed work.

The idea of using low-dimensional vectors as latent factors has found widespread use in recommendation systems.
The task of suggesting products/items to users is traditionally viewed as matrix completion where the sparse rating matrix with users as rows and items as columns is to be completed with predicted ratings. 
\citet{sarwar2000application} show how Singular Value Decomposition (SVD) can be used to decompose the rating matrix into low rank feature matrices to reduce dimensions of the rating matrix. 
This gave rise to the widely used matrix factorization techniques for predicting ratings~\cite{koren2009matrix} in which the user and item factors capture the similarities amongst them. 
Conventional matrix factorization techniques predict ratings directly as the dot product of the factors of the user and the item, and use regularized least-squares as the loss function to optimize. 
Our model here however uses the \emph{probabilistic} interpretation of matrix factorization~\cite{salakhutdinov08:probabilistic} and uses the sigmoid function with log-likelihood as it is a generalization of PCA to binary matrices~\cite{collins2001generalization}.

Since many collaborative filtering applications often have auxiliary information available for users and products/businesses, a number of approaches have studied how this information can be combined with matrix factorization for better rating prediction.
If the auxiliary observation can be treated as fully-observed and noise-free, it can be used in conjunction with the neighborhood model to augment the matrix factorization objective~\cite{koren08kdd}.
In practice, however, the auxiliary data is commonly noisy and incomplete, and thus has to be explicitly modeled for adequately leveraging it.
\citet{mcauley2012learning} combines matrix factorization with review text by modeling the words using a LDA topic model, and aligning the item/user latent vector with the review text topic vectors to learn better factors for rating prediction. 
\citet{ling2014ratings} similarly combine review text, but use mixture of Gaussian instead of matrix factorization, avoiding any transformation of the factors and thus retaining the interpretability of latent topics.
\citet{ganu2009beyond} predict ratings for restaurants from the review text alone, but require additional manually labeled data for classifying the sentiment and aspects of sentences. 
External information in the form of item taxonomies have also been investigated, for example, \citet{weng2008exploiting} combine users' preferences with the item types to learn type-level preferences, while also addressing the cold-start problem for items with only taxonomic information, but do not employ any factorization model. 
\citet{koenigstein2011yahoo} use global item biases in the Yahoo! Music dataset by using shared parameters amongst items with a common ancestor in the taxonomy hierarchy. 
\citet{koren2010collaborative} incorporates temporal dynamics into matrix factorization to learn changes in user movie preferences that occur over time, whereas, \citet{mcauley2013amateurs} argue that, to enjoy certain kinds of products such as \emph{beer} and \emph{gourmet foods}, one requires a certain level of \emph{expertise}, hence their model tries to combine temporal ratings data to make better personalized recommendations according to the \emph{experience} of each user.
Methods described above propose models that are specific to their domains, and thus the generalization capabilities of these models is unclear.

An alternative approach is to combine all the data and represent it using tensors, allowing the use of tensor factorization, an extension of matrix factorization to tensors.
For example, the approach by \citet{de2000multilinear} is used to predict tags for a user-item pairs~\cite{symeonidis2008tag,rendle2010pairwise} and to predict user ratings by integrating context information as a tensor \cite{karatzoglou2010multiverse}. 
The main shortcoming of such approaches, however, is that they model only a single additional source of information, and further, focus on predicting only the relation of interest. 

To model multiple relations in a joint manner, collective matrix factorization~\cite{singh2008relational} extend the idea of matrix factorization to multiple matrices.
The rows and columns of the matrices have corresponding latent factors, with shared latent factors for entities that appear as rows or columns in multiple matrices.
These approaches learn parameters for entities by jointly factorizing all of these matrices, and thus learn factors that predict multiple matrices.
The empirical evaluation on relatively small databases with only two relations did not show considerable improvements; this is expected since collective factorization requires, and would benefit from, larger datasets.
We use this model with the logistic/sigmoid formulation in this paper, combined with stochastic gradient descent (SGD) for optimization, and evaluate on $26$ large-scale, multi-relation real-world datasets from Amazon and Yelp, combined.

Our formulation of relational data, and the collective factorization model, can be easily extended. 
For example, the current formulation assumes at most a single relation exists between any specific pair of entities (since $P_\Phi$ is independent of relation $r$ in Eq~\ref{eq:model}).
Although this assumption holds for many applications, we can extend this model to multiple relations between the same pair of entities by introducing latent factors for the relations, similar to CP-decomposition (or PARAFAC) and recently proposed RESCAL~\cite{nickel2011three}. 
\citet{krompass2014querying} obtains highly compressed representations of large triple stores by using RESCAL to represent them as Probabilistic Databases (PDBs) and presents methods to efficiently answer complex queries on PDBs by breaking them into sub-queries.

Our model also assumes binary absence/presence relations, however non-Boolean binary relations can be modeled either by treating them as multiple binary relations, or by using a different function than the sigmoid to predict the value of the relation, while $n$-ary relations can be modeled as tensors with CP decomposition. 
It is worth mentioning that the Yelp dataset contains such deviations from our assumptions: the business attributes are discrete valued (Wi-Fi: Free, Paid, No) which is converted to multiple Boolean yes/no entities (Wi-Fi:Free, Wi-Fi:Paid, Wi-Fi:No), while the reviews in both Amazon and Yelp are $3$-way relation between users, businesses, and words which we split into two binary relations.



\newpage
\section{Conclusions and Future Work}

In this paper, we presented the application of the collective relational factorization model for improving user preference prediction.
By learning entity embeddings that are shared between all the relations the entity participates in, the model is able to combine multiple sources of evidence, predicts relations of multiple types, and further, allows computation of similarity between entities that do not share any direct relations. 
We presented empirical evaluation of user preference prediction that demonstrates that the collective model achieves higher accuracy with access to additional evidence.
We also investigated \emph{cold-start} evaluation for businesses, and showed that the collective model is accurate in predicting ratings (and attributes) even when none of the ratings (and attributes, respectively) of the business have been observed.
We additionally explore joint visualization of categories, business attributes, and review words, facilitated by the collective factors.
The code for the algorithm, along with data processing and evaluation, is available for download\footnote{\url{http://nitishgupta.github.io/factorDB/}}.

We would like to explore a number of avenues for future work.
As we described in \S~\ref{sec:related}, we will extend our collective factorization representation of relational data to support $n$-ary relations (by using tensor factorization) and to non-binary, multi-valued relations (for example, by introducing additional factors for relations).
These extensions will enable us to support a wider variety of relations and databases; we will, for example, be able to model the complete Yelp schema, including attributes such as tips, locations, temporal information, and review tags, with a single collective factorization model.
We will also investigate applications of this model on relational databases from other domains.

%
\footnotesize
\bibliographystyle{abbrvnat}
\bibliography{sigproc}  

\end{document}